\documentclass{article}
\usepackage[final]{neurips_2026}
\makeatletter
\providecommand{\@trackname}{}
\makeatother
\usepackage[utf8]{inputenc}
\usepackage[T1]{fontenc}
\usepackage{hyperref}
\usepackage{url}
\usepackage{booktabs}
\usepackage{amsfonts}
\usepackage{amsmath,amssymb}
\usepackage{nicefrac}
\usepackage{microtype}
\usepackage{xcolor}
\usepackage{array}
\usepackage{longtable}
\usepackage{graphicx}
\usepackage{subcaption}
\usepackage{multirow}
\usepackage{makecell}
\usepackage{xcolor}
\usepackage{tcolorbox}[enhanced]
\newcommand{\blfootnote}[1]{%
  \begingroup
  \renewcommand{\thefootnote}{}%
  \footnotetext{#1}%
  \endgroup
}
\tcbuselibrary{breakable, listings, skins}

\newtcblisting{promptbox}[1]{
  breakable,
  enhanced,
  colback=gray!3,
  colframe=gray!45,
  title=#1,
  fonttitle=\bfseries,
  boxrule=0.5pt,
  arc=2pt,
  left=6pt,
  right=6pt,
  top=6pt,
  bottom=6pt,
  listing only,
  listing options={
    basicstyle=\ttfamily\scriptsize,
    breaklines=true,
    breakatwhitespace=true,
    columns=fullflexible,
    keepspaces=true,
    showstringspaces=false,
    tabsize=2
  }
}
\title{CanvasAgent: Enabling Complex Image Creation and Editing via Visual Tool Orchestration}

\author{%
  {\normalfont\small
  Hairui Zhu$^{1}$ \quad
  Yiying Yang$^{1}$ \quad
  Tengjin Weng$^{1}$ \quad
  Ziyu Lu$^{1}$} \\
  {\normalfont\mdseries\small
  Xiao Yao$^{1}$ \quad
  Xiaoyang Ye$^{1}$ \quad
  Lin Ma \quad
  Wenhao Jiang$^{1}$\thanks{Corresponding author.}} \\
  \\[-0.2em]
  {\normalfont\mdseries\footnotesize
  $^{1}$Guangdong Laboratory of Artificial Intelligence and Digital Economy (SZ)} \\
  {\normalfont\mdseries\footnotesize
  Shenzhen, China}
}

\begin{document}

\maketitle
\blfootnote{
Code and datasets are available at:
\url{https://github.com/GML-FMGroup/CanvasAgent}\\
\url{https://huggingface.co/datasets/GML-FMGroup/CanvasCraftSFT}\\
\url{https://huggingface.co/datasets/GML-FMGroup/CanvasCraftRL}.
}
\begin{abstract}

Complex image creation and editing often require more than a single generation
or editing model. A user request may involve synthesizing images, localizing
objects, segmenting regions, editing selected content, compositing intermediate
assets, reading text, and enhancing the final result. Such tasks shift
multimodal agents from perception-augmented reasoning to manipulation-centered
visual creation, where tools must actively transform visual states rather than
merely inspect them. However, existing multimodal tool-use agents are mostly
optimized for perception, search, or domain-specific editing, and lack
large-scale supervision for executable image-creation trajectories. In this paper, we introduce \textbf{CanvasCraft}, a large-scale multimodal
tool-use dataset for complex image creation and editing, and
\textbf{CanvasAgent}, a tool-augmented multimodal agent that learns to
orchestrate heterogeneous visual tools through multi-turn interaction.
CanvasCraft contains $140K$ fully annotated executable trajectories and $10K$
 RL task specifications. CanvasAgent is first trained with SFT to
learn executable reasoning-action trajectories, and is then optimized with GRPO
using a hybrid reward that combines outcome- and process-level signals. During
rollout, CanvasAgent inspects intermediate results, tracks visual assets, and
adapts tool decisions to the evolving visual state. Experiments evaluate both
final image quality and trajectory behavior, demonstrating the effectiveness of
\textbf{CanvasAgent} and the proposed dataset for complex multi-tool image
creation workflows.

\end{abstract}

\section{Introduction}

Image creation and editing have advanced rapidly with diffusion models,
instruction-guided editors, and multimodal large language models
(MLLMs)~\cite{rombach2022latent,brooks2023instructpix2pix,saharia2022sr3}.
Yet many practical requests still exceed a single model call. A user may ask
an assistant to generate a scene, locate and segment an object, replace only
that region, add text or another object, crop the result, and enhance its
resolution. Such requests require generation, localization, segmentation,
editing, compositing, OCR, geometric transformation, and enhancement to be
coordinated within one coherent workflow.

These workflows differ from standard image generation or single-step editing in
three key ways. First, they are long-horizon: later operations depend on visual
artifacts produced by earlier tools. Second, they are visually grounded: after
each tool call, the agent must inspect the intermediate output instead of
assuming success. Third, they are stateful: multiple images, masks, crops,
extracted objects, and edited variants may coexist, and the agent must select
the correct asset for each subsequent operation. These properties make complex
image creation and editing a trajectory-learning problem rather than a simple
prompt-to-image or instruction-to-image task.

Existing multimodal tool-use research addresses parts of this problem but not
the full setting. Early tool-augmented systems connect language models with
visual foundation models or external experts for multi-step reasoning and
editing~\cite{wu2023visualchatgpt,yang2023mmreact,shen2023hugginggpt,
gupta2023visualprogramming,surismenon2023vipergpt}. Recent agentic MLLMs
improve active visual perception, search, and executable
reasoning~\cite{zheng2025deepeyes,hong2025deepeyesv2,zhang2025thyme,
su2025pixelreasoner,zhao2025pyvision}, but their tasks mainly target
understanding, search, or reasoning. Image-editing systems and datasets,
including instruction-guided editing and photo-retouching
agents~\cite{brooks2023instructpix2pix,zhang2023magicbrush,fu2023mgie,
sheynin2023emuedit,bai2025qwenimage,lin2025jarvisart,lin2025jarvisevo}, are
closer to our target domain, yet typically focus on single-model editing or
retouching in specialized environments. They do not provide large-scale
executable trajectories that combine heterogeneous tools, intermediate visual
observations, and explicit multi-asset state.

To address this gap, we introduce \textbf{CanvasCraft}, a large-scale
multimodal tool-use dataset for complex image creation and editing workflows.
CanvasCraft contains two complementary subsets. \textbf{CanvasCraft-SFT}
provides fully annotated execution trajectories with user instructions,
optional input images, step-level reasoning, tool calls, parameters, outputs,
intermediate visual artifacts, and final images. \textbf{CanvasCraft-RL}
provides task specifications with expected tool sets, enabling reinforcement
learning to explore tool ordering, parameterization, recovery, and stopping
strategies without imitating a fixed trajectory.

We instantiate \textbf{CanvasAgent}, a tool-augmented MLLM
trained in two stages. Supervised fine-tuning on CanvasCraft-SFT provides an
initialization for valid tool invocation and cross-tool dependencies.
Reinforcement learning on CanvasCraft-RL then refines the policy with GRPO and
a hybrid trajectory-level reward combining final-output alignment, visual
quality, reasoning validity, rule-based executability, and efficiency
penalties. During execution, CanvasAgent follows a visual-first protocol and
maintains explicit image-asset references, allowing it to revise plans, switch
tools, or roll back to earlier outputs.

Our contributions are summarized as follows:
\begin{itemize}

    \item We introduce \textbf{CanvasAgent}, the first tool-augmented multimodal
    agent designed for complex image creation and editing. CanvasAgent moves
    beyond passive visual perception by actively orchestrating heterogeneous
    visual tools for generation, editing, extraction, composition,
    transformation, and enhancement through multi-turn reasoning.

    \item We construct \textbf{CanvasCraft}, the first large-scale multimodal
    tool-use dataset for complex image creation and editing. CanvasCraft covers
    diverse creation scenarios, tool combinations, and multi-turn visual
    workflows, with \textbf{CanvasCraft-SFT} providing fully annotated
    multi-step execution trajectories and \textbf{CanvasCraft-RL} providing
    diverse and challenging task specifications for reinforcement learning.

    \item We design a task-specific \textbf{hybrid reward} for complex visual
    creation and integrate it into a two-stage SFT+GRPO training framework.
    The reward combines LLM-as-judge signals for image-prompt alignment,
    aesthetic quality, and trajectory validity with rule-based process checks
    and efficiency penalties, enabling robust optimization of final images and
    tool-use processes.

\end{itemize}

\section{Related Work}
\vspace{-1mm}

\paragraph{Tool-Augmented Multimodal Agents}

Tool use has become a central mechanism for extending language and multimodal
models beyond direct generation. ReAct~\cite{yao2023react} formalizes the
reason-action-observation pattern, and early multimodal systems connect
language models with visual foundation models or expert modules for planning,
execution, and programmatic visual reasoning~\cite{wu2023visualchatgpt,
yang2023mmreact,shen2023hugginggpt,gupta2023visualprogramming,
surismenon2023vipergpt,lu2023chameleon}. Recent agentic MLLMs further develop
active visual inspection, search, and executable reasoning through operations
such as cropping, zooming, Python execution, and heterogeneous tool
use~\cite{zheng2025deepeyes,hong2025deepeyesv2,zhang2025thyme,
su2025pixelreasoner,shen2025zoomeye,zhao2025pyvision,song2025codedance,
chng2025sensenovamars,zhang2025skyworkr1v4}. These works mainly target
perception, visual search, reasoning, or general multimodal assistance. Our
work instead focuses on complex image creation and editing workflows, where
the agent must produce a final visual artifact by coordinating generation,
localization, editing, compositing, OCR, and enhancement tools across stateful
trajectories.

\paragraph{Image Editing and Creation Workflows}

Image generation and editing models have made substantial progress on
individual operations. Latent diffusion models enable high-quality
synthesis~\cite{rombach2022latent}, and instruction-guided editors improve
controllable editing, generation, and text rendering~\cite{
brooks2023instructpix2pix,zhang2023magicbrush,fu2023mgie,sheynin2023emuedit,
bai2025qwenimage}. These models are strong building blocks, but they typically
execute a single instruction in one model call and do not explicitly manage
multi-step tool dependencies or multiple intermediate visual assets.

Photo-retouching agents are closer to our setting because they coordinate
editing operations over multiple steps. JarvisArt~\cite{lin2025jarvisart}
controls Lightroom operations with an MLLM agent, and
JarvisEvo~\cite{lin2025jarvisevo} studies a self-evolving
edit-evaluate-reflect loop. However, their environment is primarily designed
for photo retouching. Our setting targets open-ended image creation and
editing workflows that may require generation, grounding, segmentation, object
extraction, compositing, cropping, OCR, geometric transformation, and
super-resolution within the same trajectory.

\paragraph{Learning Tool-Orchestration Policies}

Learning effective tool use requires more than exposing a model to tool
descriptions. Supervised fine-tuning can teach invocation schemas and
reasoning-action formats, but imitation alone may overfit to static
demonstrations and fail to discover better long-horizon strategies.
Reinforcement learning has therefore been used to optimize tool-use policies,
from PPO~\cite{schulman2017ppo} and GRPO~\cite{shao2024deepseekmath} to recent
search, visual reasoning, and tool-use systems~\cite{jin2025searchr1,
wu2025mmsearchr1,zhang2025openthinkimg,chen2025toolr1,li2025toolscope,
liu2025visualtoolbench,wang2025adatoolerv,yan2026metis}.

Complex image creation and editing pose a different optimization problem. The
agent must select tools, set tool parameters, track intermediate assets, and
decide whether the current visual result is sufficient. Reward design must
therefore evaluate both the final output image and the execution process.
We address this with a two-stage SFT+GRPO framework and a hybrid reward that
jointly optimizes visual reasoning, image quality, trajectory validity, and
robust tool-use behavior.


\begin{figure}[t]
    \centering
    \includegraphics[width=0.98\linewidth]{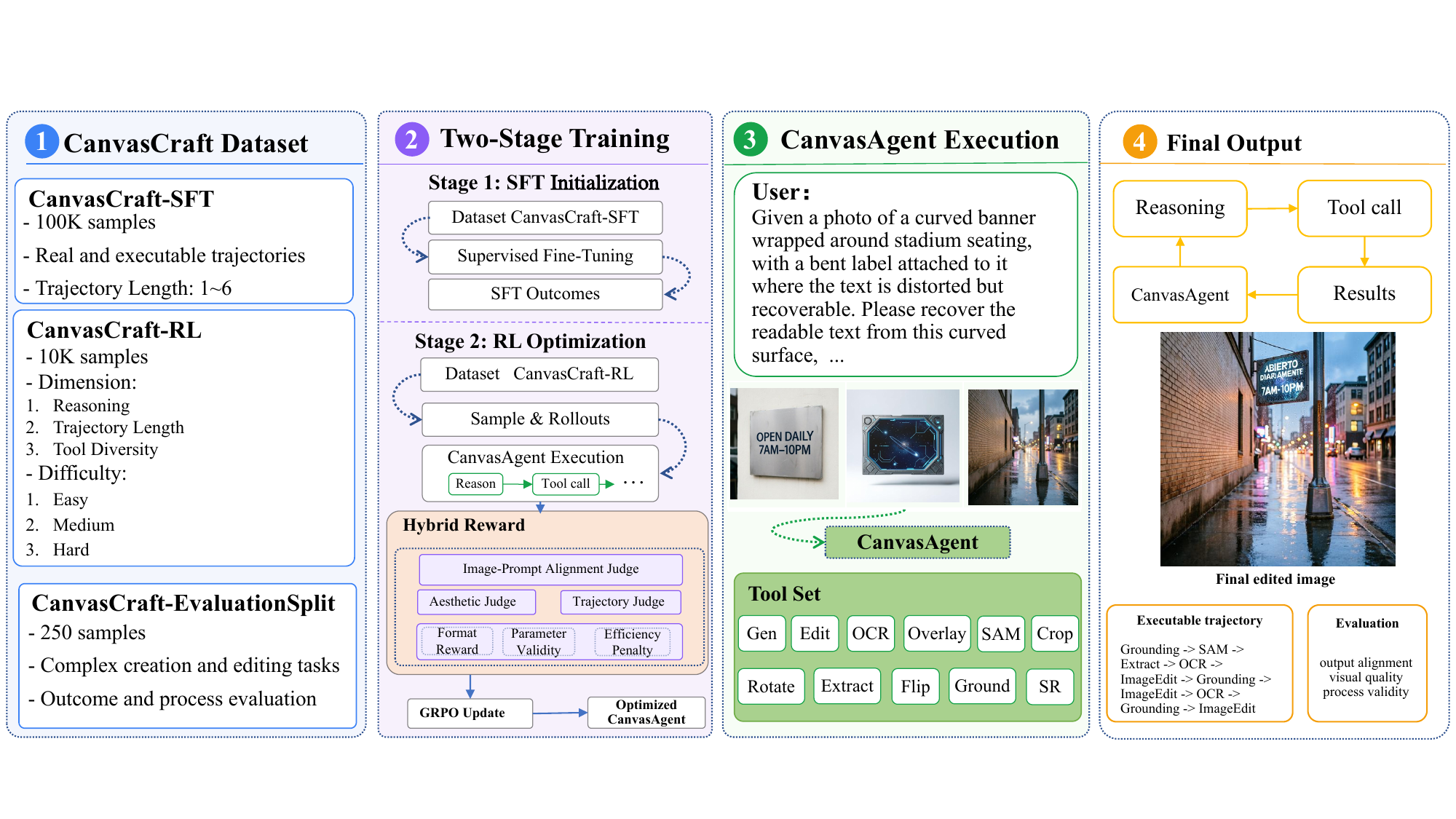}
    \vspace{-8mm}
    \caption{
Overview of CanvasCraft and CanvasAgent. CanvasCraft provides supervised
tool-use trajectories and RL task specifications for training CanvasAgent,
which learns to orchestrate visual tools for complex image creation and
editing.
}
    \label{fig:overall-framework}
\vspace{-5mm}
\end{figure}

\vspace{-1mm}
\section{Method}
\vspace{-2mm}

This section describes the data construction and training framework of
CanvasAgent, illustrated in Fig.~\ref{fig:overall-framework}. We first construct
CanvasCraft, a large-scale multimodal tool-use dataset for complex image
creation and editing. It contains CanvasCraft-SFT, which provides fully
annotated multi-step trajectories, and CanvasCraft-RL, which provides diverse
task-level specifications for RL training. We then train CanvasAgent with a
two-stage SFT+RL framework: supervised fine-tuning bootstraps stable tool-use
behavior from expert trajectories, while reinforcement learning with  GRPO improves long-horizon planning
with a task-specific hybrid reward.
CanvasAgent operates with 11 specialized visual tools covering generation,
editing, localization, segmentation, extraction, compositing, geometric
transformation, OCR, and super-resolution. During execution, it perceives the
current visual state, reasons over intermediate assets, and plans subsequent
tool calls, enabling adaptive multi-tool orchestration guided by user
instructions and evolving visual feedback.

\subsection{CanvasCraft Dataset and Construction}
\label{sec:CanvasCraft}

To train CanvasAgent for complex image creation and editing, we construct
CanvasCraft around two principles: coverage and diversity. Coverage ensures
that the data spans essential visual operations, while diversity encourages
variation in reasoning difficulty, trajectory length, and tool combination. 
CanvasCraft contains two
complementary subsets. CanvasCraft-SFT provides executable multi-step
trajectories for supervised tool-use learning, whereas CanvasCraft-RL provides
task-level specifications for RL-based policy exploration.





\vspace{-1mm}
\subsubsection{Dataset Construction}
\vspace{-1mm}

\paragraph{CanvasCraft Tool Set}

\begin{table*}[h]
\centering
\small
\caption{Unified visual tool set used in CanvasCraft and CanvasAgent. Each tool exposes structured inputs and outputs and is backed by a concrete visual model or script.}
\label{tab:toolset}
\setlength{\tabcolsep}{3.0pt}
\renewcommand{\arraystretch}{1.12}

\begin{tabular}{@{}p{0.15\textwidth}p{0.21\textwidth}p{0.22\textwidth}p{0.18\textwidth}p{0.18\textwidth}@{}}
\toprule
\textbf{Tool} & \textbf{Function} & \textbf{Key Inputs} & \textbf{Output} & \textbf{Backend} \\
\midrule
\texttt{Generation}
& Text-to-image synthesis
& \texttt{prompt}
& Generated image
& FLUX.2-Klein-4B \\

\texttt{Edit}
& Instruction-based image editing
& \texttt{image}, \texttt{edit prompt}, optional \texttt{mask}
& Edited image
& FLUX.2-Klein-4B \\

\texttt{Grounding}
& Object localization
& \texttt{image}, \texttt{reference text}
& Bounding box
& Grounding-DINO \\

\texttt{SAM}
& Mask generation
& \texttt{image}, \texttt{bounding box}
& Segmentation mask
& SAM \\

\texttt{Extract}
& Object extraction
& \texttt{image}, \texttt{SAM mask}
& Extracted object
& Python script \\

\texttt{Overlay}
& Object/text compositing
& \texttt{base image}, \texttt{overlay type}, \texttt{content}, \texttt{position}
& Composited image
& Python script \\

\texttt{Crop}
& Region cropping
& \texttt{image}, \texttt{bounding box}
& Cropped image
& Python script \\

\texttt{OCR}
& Text recognition
& \texttt{image}
& Recognized text
& Paddle-OCR \\

\texttt{Rotate}
& Orientation correction
& \texttt{image}, \texttt{angle}
& Rotated image
& Python script \\

\texttt{Flip}
& Horizontal mirroring
& \texttt{image}
& Flipped image
& Python script \\

\texttt{SR}
& Super-resolution
& \texttt{image}
& $4\times$ enhanced image
& Real-ESRGAN \\
\bottomrule
\end{tabular}
\end{table*}
CanvasCraft is built on a unified toolkit of 11 heterogeneous visual tools,
including generation, editing, grounding, segmentation, extraction,
compositing, cropping, OCR, rotation, flipping, and super-resolution. The full
tool specification is provided in Table~\ref{tab:toolset}. Each tool is
implemented as a low-level operation with structured JSON-style schema, enabling flexible composition into multi-step visual workflows.
Because visual tool use depends on intermediate image states, CanvasCraft
emphasizes tool chaining, asset management, and long-horizon workflow
execution.

\begin{figure}[t]
    \centering
    \includegraphics[width=\linewidth]{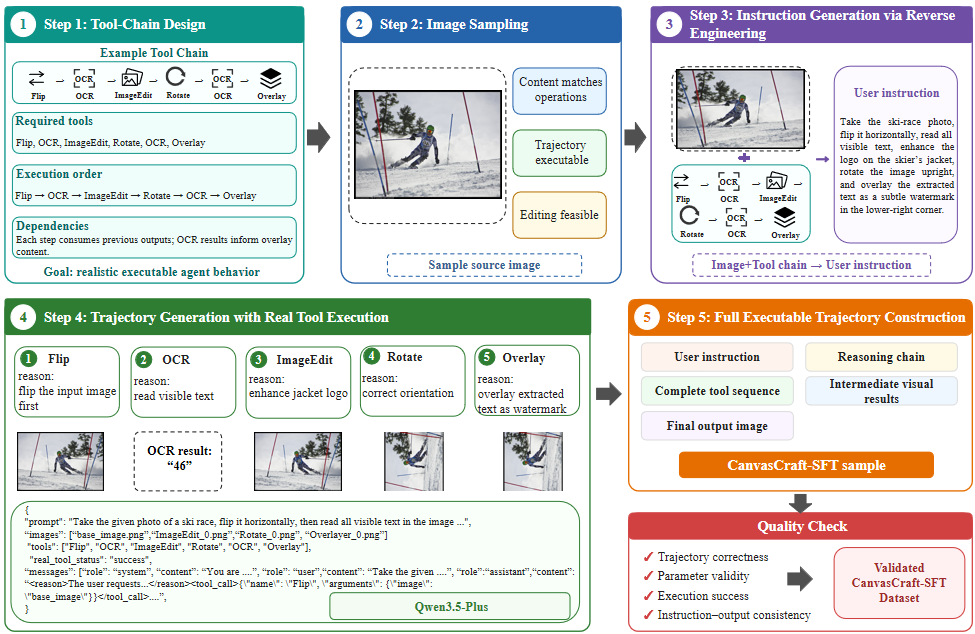}
    \caption{
    CanvasCraft-SFT data construction pipeline. The pipeline constructs executable
tool-use trajectories through tool-chain design, image sampling, instruction
generation via reverse engineering, and real tool execution with quality control.
    }
    \label{fig:sft_data_pipeline}
\end{figure}
\paragraph{Construction of CanvasCraft-SFT.}


CanvasCraft-SFT provides executable trajectory supervision for visual tool orchestration. Instead of collecting only input--output image pairs, each sample records how a user instruction is decomposed and completed through real tool execution, including step-level reasoning, tool calls, structured parameters, intermediate assets, and final outputs. This allows the Agent to learn valid tool invocation, asset referencing, and cross-tool dependencies in multi-step visual workflows. Formally, each sample is denoted as $d_{\mathrm{SFT}}=(Q,I,\tau,I_{\mathrm{final}})$, where $Q$ is the user instruction, $I$ denotes the optional input image set, $\tau$ is the executable tool-use trajectory, and $I_{\mathrm{final}}$ is the final output image.

As shown in Fig.~\ref{fig:sft_data_pipeline}, we construct CanvasCraft-SFT through a tool-chain-driven pipeline. We first design tool-chain templates covering atomic visual operations and representative inter-tool dependencies. For each template, we select appropriate input images sampled from PICO-Banana-400K~\cite{qian2025picobanana}, and reverse-engineer a natural user instruction from the intended tool-use behavior. The instruction is then executed in the tool environment to produce reasoning traces, JSON-style tool calls, intermediate assets, and final images. We retain only trajectories that pass quality control, including checks for tool-call parsability, parameter validity, asset-reference consistency, execution success, and redundancy.

\begin{figure}[t]
    \centering
    \includegraphics[width=1.05\linewidth]{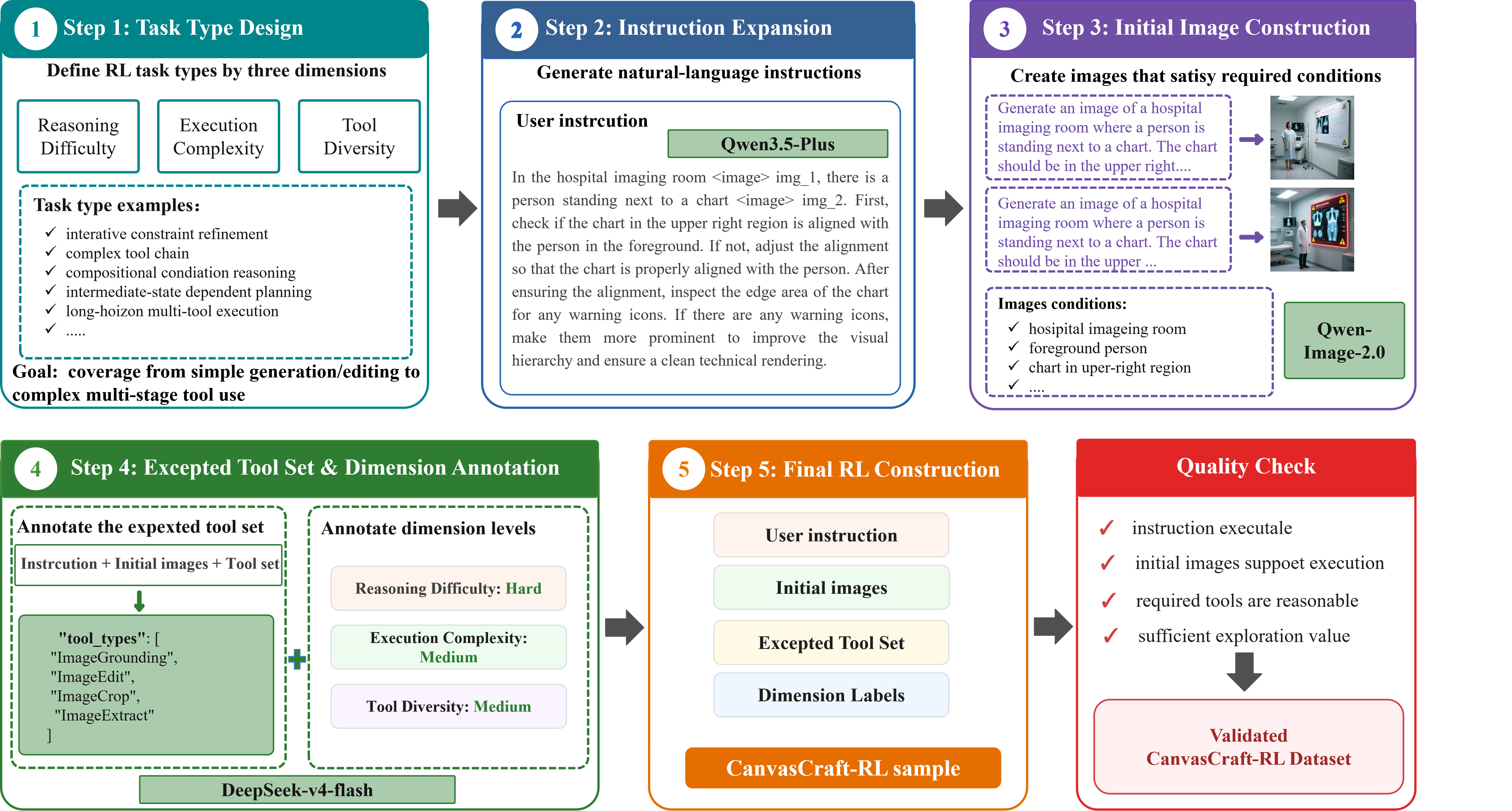}
    \caption{
    CanvasCraft-RL data construction pipeline. The pipeline generates difficulty-aware
tasks by expanding task seeds into user instructions, constructing
aligned initial images when needed, annotating expected tool sets, and filtering
ambiguous or low-quality instances for RL training.
    }
    \label{fig:rl_data_pipeline}
\end{figure}
\paragraph{Construction of CanvasCraft-RL.}


Unlike CanvasCraft-SFT, CanvasCraft-RL only contains the expected tool set serves  as weak supervision, allowing the agent to explore alternative tool ordering, parameterization, verification, and stopping strategies during rollout. Each sample is represented as $d_{\mathrm{RL}}=(Q,\mathcal{I}_0,\mathcal{T}^{})$, where $Q$ is the user instruction, $\mathcal{I}_0$ denotes the optional input image set, and $\mathcal{T}^{}$ is the expected tool set for the task. 

As shown in Fig.~\ref{fig:rl_data_pipeline}, CanvasCraft-RL is constructed around three task dimensions: \textbf{Reasoning (R)}, \textbf{Trajectory Length (L)}, and \textbf{Tool Diversity (D)}. These dimensions describe the reasoning complexity, expected execution length, and tool heterogeneity of each task, respectively. We first generate diverse task seeds to cover these three dimensions, and then expand each seed into a natural user instruction with an expected tool set $\mathcal{T}^{*}$. After task generation, each sample is characterized along the R/L/D dimensions and assigned difficulty levels according to the criteria in Table~\ref{tab:rct_definition}. The generated tasks are then filtered by DeepSeek-V4-Flash and further verified by human annotators to remove ambiguous, overly simple, or R/L/D-inconsistent instances. For visually grounded tasks, we generate one or more initial images aligned with the instruction.

Together, CanvasCraft-SFT and CanvasCraft-RL provide complementary supervision: the former teaches executable tool-use patterns through complete trajectories, while the latter enables reinforcement learning to optimize flexible planning and dynamic multi-tool orchestration under task-level weak supervision.
\begin{table}[t]
\centering
\small
\setlength{\tabcolsep}{4pt}
\renewcommand{\arraystretch}{1.15}
\caption{Difficulty definitions for the 10K CanvasCraft-RL tasks across Reasoning (R), Trajectory Length (L), and Tool Diversity (D).}
\label{tab:rct_definition}
\begin{tabular}{p{2.4cm} p{3.3cm} p{3.5cm} p{3.5cm}}
\toprule
\textbf{Dimension} & \textbf{Easy} & \textbf{Medium} & \textbf{Hard} \\
\midrule

\textbf{Reasoning (R)} &
Single execution path with weak conditional reasoning or minimal state awareness &
Multi-condition composition with local path selection based on intermediate results &
Nested conditions, strong state dependency, and significantly branching execution paths \\
\midrule
\textbf{Trajectory Length (L)} &
1--4 execution steps with short tool chains &
5--8 steps with explicit task decomposition or serial tool interaction &
9+ steps with long-horizon execution and multi-stage coordination \\
\midrule
\textbf{Tool Diversity (D)} &
1--2 distinct tools with simple composition &
3--4 distinct tools with stable collaborative patterns &
5+ distinct tools requiring heterogeneous tool orchestration and dynamic switching \\

\bottomrule
\end{tabular}
\end{table}

\begin{figure}[h]
    \centering
    \includegraphics[width=\linewidth]{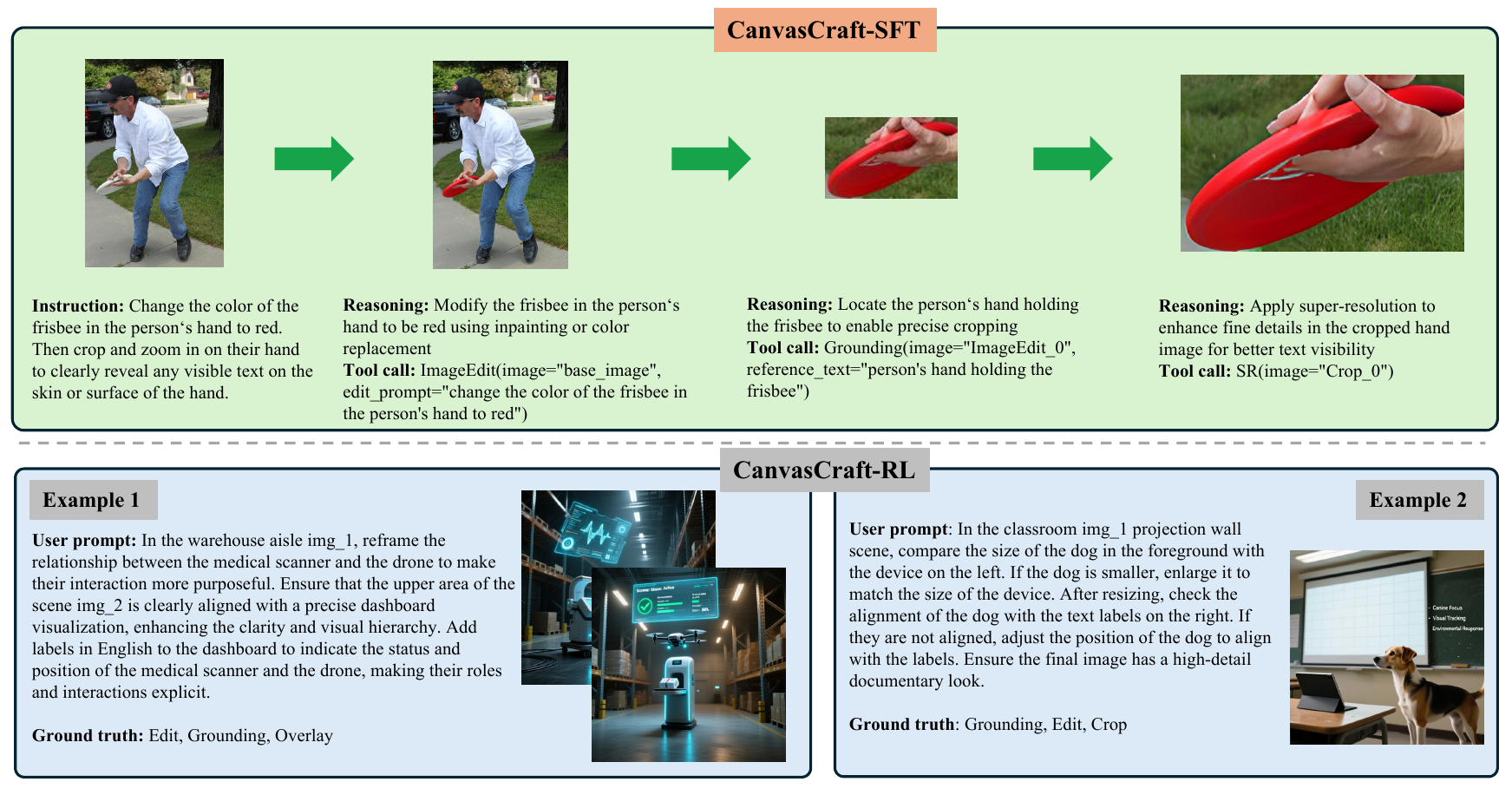}
    \caption{
    CanvasCraft data example. CanvasCraft-SFT provides the complete execution
    chain, including step-level reasoning and tool calls. CanvasCraft-RL
    provides task specifications without complete trajectories, requiring agents
    to explore tool ordering, parameterization, and intermediate operations
    during rollout.
    }
    \label{fig:data_example}
\end{figure}



\subsubsection{Dataset Composition}
CanvasCraft contains \textbf{140K} SFT trajectories, \textbf{10K} RL task
specifications, and a manually curated \textbf{250}-sample evaluation
benchmark. A representative data instance is shown in  Fig.~\ref{fig:data_example}. The dataset covers diverse visual tools,
tool-chain structures, and task difficulties, supporting both supervised
tool-use bootstrapping and RL-based policy optimization.

\textbf{CanvasCraft-SFT} providing broad coverage of all 11 visual tools and diverse executable supervision through fully annotated tool-use trajectories. The trajectories range from single-tool operations to short dependency chains and high-complexity multi-turn workflows. In particular, 3.8K examples involve longer execution paths, richer tool combinations, and more involved inter-tool dependencies, enabling the model to acquire valid tool invocation formats, basic tool dependencies, and initial long-horizon execution patterns.
\begin{table}[h]
\centering
\small
\setlength{\tabcolsep}{10pt}
\renewcommand{\arraystretch}{1.15}
\caption{Difficulty distribution of the 10K CanvasCraft-RL tasks across the R/L/D dimensions.}
\label{tab:rct_distribution}
\begin{tabular}{lccc}
\toprule
\textbf{Dimension} & \textbf{Easy} & \textbf{Medium} & \textbf{Hard} \\
\midrule

\textbf{Reasoning (R)} & $761$ & $1,807$ & $7,432$ \\

\textbf{Trajectory Length (L)} & $476$ & $4,687$ & $4,837$ \\

\textbf{Tool Diversity (D)} & $2,317$ & $5,631$ & $2,052$ \\

\bottomrule
\end{tabular}
\end{table}

\textbf{CanvasCraft-RL} is organized along reasoning difficulty, trajectory length, and tool diversity (R/L/D), with an emphasis on medium and hard levels; the full distribution is reported in Table~\ref{tab:rct_distribution}. Unlike CanvasCraft-SFT, this subset only provides task-level supervision rather than fully annotated trajectories, encouraging the agent to learn task decomposition, state-dependent planning, and heterogeneous tool orchestration during reinforcement learning.

\subsection{Two-Stage Training Framework}

\label{sec:training_framework}

Based on the two subsets of CanvasCraft, we train CanvasAgent with a two-stage
SFT+GRPO framework as shown in Fig.~\ref{fig:overall-framework}. SFT bootstraps stable tool-use behavior from complete
executable trajectories, while GRPO optimizes rollout-level multi-tool planning
under weak supervision. This design combines the reliability of trajectory
imitation with the flexibility of reinforcement learning-based policy
exploration.


\subsubsection{SFT for Tool-Use Bootstrapping}\label{sec:sft}


In the SFT stage, CanvasAgent is trained on CanvasCraft-SFT with the standard
next-token prediction objective. Given complete expert trajectories, the model
learns valid reasoning-action formats, JSON-style tool calls, parameter
generation, cross-tool dependencies, and references to intermediate visual
assets. This stage stabilizes subsequent reinforcement learning, since direct
RL over executable visual tools often leads to invalid tool calls, unstable
rollouts, and sparse rewards. Nevertheless, SFT is limited to imitating
demonstrated trajectories and cannot sufficiently explore alternative tool-use
strategies, motivating the RL stage.

\subsubsection{RL for Image Planning and Manipulation}
\label{sec:rl}



After SFT initialization, we further optimize CanvasAgent on CanvasCraft-RL with
group relative policy optimization (GRPO)~\cite{shao2024deepseekmath}. For each
task, GRPO samples multiple executable rollouts and updates the policy according
to their relative rewards, without requiring an additional value model. This
stage encourages the agent to explore different task decompositions and
tool-use strategies under weak supervision. The optimization signal comes from
our task-specific hybrid reward, which evaluates both final visual quality and
tool-use process validity.




\vspace{-1mm}
\subsection{Hybrid Reward Design}
\vspace{-1mm}

Diverse image creation and manipulation tasks are difficult to evaluate because
they involve multiple intermediate steps, diverse tool choices, and subjective
visual quality. To guide RL training, we design a \textbf{task-specific hybrid
reward} that evaluates both the final output and the execution trajectory:

\begin{equation}
\begin{aligned}
R(\tau) &=
\underbrace{0.3 \cdot R_{\mathrm{align}}(\tau)
+ 0.1 \cdot R_{\mathrm{aes}}}_{\text{Outcome Score}}(\tau)
+ \underbrace{0.2 \cdot R_{\mathrm{traj}}(\tau)
+ 0.4 \cdot R_{\mathrm{rule}}}_{\text{Process Score}}(\tau).
\end{aligned}
\end{equation}


\subsubsection{Outcome Score}
Outcome scores are computed with expert LLM-as-judge evaluators.

\paragraph{Alignment Score.}
$R_{\mathrm{align}}$ evaluates whether the final image satisfies the user's
instruction. Given the image-prompt pair and the final image, it focuses on
requested objects, attributes, actions, spatial relations, colors, text,
quantities, and edited regions.

\paragraph{Aesthetic Score.}
$R_{\mathrm{aes}}$ evaluates the perceptual quality of the final image. It
takes the final image as input and scores composition, color, lighting,
clarity, texture naturalness, style consistency, artifacts, and blur. This
score is independent of instruction alignment and trajectory quality.

These judges capture semantic fidelity and visual quality, which cannot be
fully measured by deterministic rule-based signals alone.

\subsubsection{Process Score}

\paragraph{Trajectory Score.}
$R_{\mathrm{traj}}$ evaluates whether the trajectory is reasonable for the
task. It takes the whole trajectory and the expected tool set as input. The
judge focuses on process quality, including whether the agent decomposes the
task properly, selects appropriate tools, follows valid tool dependencies, and
avoids irrelevant or unnecessary operations. It does not directly judge the
final image.

\paragraph{Rule-based Reward.}
While LLM judges evaluate high-level semantic, perceptual, and reasoning
quality, rule-based rewards provide deterministic feedback on low-level
executability. We define the rule-based reward as a combination of
format reward, action reward, and efficiency penalty:
\begin{equation}
R_{\mathrm{rule}}(\tau)
=
0.4 \cdot R_{\mathrm{format}}(\tau)
+
0.6 \cdot R_{\mathrm{action}}(\tau)
-
\lambda_{\mathrm{eff}} P_{\mathrm{eff}}(\tau),
\label{eq:process_reward}
\end{equation}
The format reward $R_{\mathrm{format}}$ verifies whether the model follows the
required reasoning-action protocol. It checks the presence of valid
\texttt{<reason>} and \texttt{<tool\_call>} blocks, whether tool calls can be
parsed as valid JSON, whether each turn contains at most one tool call, whether
termination is used properly, and whether the response avoids unsupported
extra formatting. This reward enforces syntactic compliance with the agent
interface

While the format reward checks well-formed syntax, the action reward evaluates
whether each tool invocation is executable under the current visual state. Given
a trajectory $\tau$ with $N$ parsed tool calls, we compute
\begin{equation}
R_{\mathrm{action}}(\tau)=\frac{1}{N}\sum_{i=1}^{N} q_i,
\label{eq:action_reward_avg}
\end{equation}
where the per-action validity score $q_i$ is defined as
\begin{equation}
q_i =
0.25 r_i^{\mathrm{name}}
+ 0.25 r_i^{\mathrm{schema}}
+ 0.20 r_i^{\mathrm{ref}}
+ 0.20 r_i^{\mathrm{spec}}
+ 0.10 r_i^{\mathrm{exec}},
\label{eq:step_quality}
\end{equation}

The five sub-scores are computed as
\begin{equation}
\begin{alignedat}{2}
r_i^{\mathrm{name}} &=
\mathbb{I}\!\left[f_i \in \mathcal{T}\right],
&\qquad
r_i^{\mathrm{schema}} &=
\mathbb{I}\!\left[\mathrm{Req}(f_i) \subseteq
\mathrm{NonEmptyArgs}(a_i)\right], \\
r_i^{\mathrm{ref}} &=
\frac{1}{|\mathrm{Refs}(a_i)|}
\sum_{v \in \mathrm{Refs}(a_i)}
\mathbb{I}\!\left[v \in \mathcal{A}_{i-1}\right],
&\qquad
r_i^{\mathrm{spec}} &=
\mathbb{I}\!\left[\mathrm{SpecCheck}(f_i,a_i,\mathcal{A}_{i-1})=1\right], \\
r_i^{\mathrm{exec}} &=
\mathbb{I}\!\left[\mathrm{Exec}(a_i)\geq 0\right].
\end{alignedat}
\label{eq:reward_components}
\end{equation}

Here, $f_i$ is the selected tool, $a_i$ denotes its structured arguments,
$\mathcal{T}$ is the set of valid tools, and $\mathcal{A}_{i-1}$ is the set of
available visual assets before step $i$. The term
$r_i^{\mathrm{name}}$ checks tool-name validity, while
$r_i^{\mathrm{schema}}$ verifies whether the required arguments of tool $f_i$
are present and non-empty. The reference score $r_i^{\mathrm{ref}}$ checks
whether all image or asset references in the arguments point to valid assets in
the current state. The tool-specific score $r_i^{\mathrm{spec}}$ encodes
operation-dependent constraints, such as valid bounding boxes for localization
and segmentation, valid masks for extraction, valid positions for compositing,
and required asset types for object overlay. Finally, $r_i^{\mathrm{exec}}$
indicates whether the corresponding tool execution succeeds.

After a successful image-producing tool call, the resulting asset identifier is
added to the asset state:
\begin{equation}
\mathcal{A}_{i}
=
\mathcal{A}_{i-1}
\cup
\{\mathrm{OutputID}(f_i,i)\}
\quad
\text{if } f_i \in \mathcal{T}_{\mathrm{img}} \text{ and } r_i^{\mathrm{exec}}=1,
\label{eq:asset_update}
\end{equation}
otherwise $\mathcal{A}_{i}=\mathcal{A}_{i-1}$. Therefore,
$R_{\mathrm{action}}$ evaluates whether each tool call is compatible with the
evolving visual asset state and executable within the tool environment.

\paragraph{Efficiency Penalty.}
The efficiency penalty discourages degenerate or unnecessarily costly
trajectories:
\begin{equation}
P_{\mathrm{eff}}(\tau)
=
P_{\mathrm{error}}(\tau)
+
P_{\mathrm{repeat}}(\tau)
+
P_{\mathrm{length}}(\tau)
+
P_{\mathrm{cost}}(\tau)
+
P_{\mathrm{miss}}(\tau),
\label{eq:efficiency_penalty}
\end{equation}
Here, $P_{\mathrm{error}}$ penalizes failed tool executions,
$P_{\mathrm{repeat}}$ penalizes adjacent repeated or near-repeated tool calls,
$P_{\mathrm{length}}$ penalizes overly long reasoning segments,
$P_{\mathrm{cost}}$ discourages excessive tool usage beyond the expected
interaction budget, and $P_{\mathrm{miss}}$ penalizes missing key expected
tools. These terms prevent the agent from improving trajectory scores by
blindly invoking more tools, and instead encourage concise, purposeful, and
task-relevant tool orchestration.


Together, the hybrid reward balances \textbf{semantic alignment, visual
aesthetics, reasoning validity, rule adherence, and execution efficiency}. By
combining semantic, perceptual, procedural, and symbolic constraints, it reduces
reward hacking and provides rich supervision for robust multi-step visual
creation.

\section{Experiments}

\subsection{Experiment Settings}
\label{sec:experimental_setup}

\paragraph{Evaluation set.}
We evaluate all models on the CanvasCraft-RL evaluation split, which contains
$250$ samples. Each sample includes a user instruction, an optional input image,
and a reference tool set used only for reward computation.

\paragraph{Compared methods.}
We compare three groups of methods. First, we evaluate general-purpose MLLMs
equipped with the CanvasCraft tool set, including
\textbf{LLaVA-OneVision-7B}, \textbf{Qwen3-VL-8B-Instruct}, and
\textbf{Qwen3-VL-32B-Instruct}. These models use the same tools but are not
trained on CanvasCraft. Second, we include \textbf{Qwen-Image-2.0}, \textbf{Wan2.7-Image}, and
\textbf{GPT-Image-2} as image-only reference models. Since they directly
generate or edit the final image without producing
reasoning--action--observation trajectories, their results serve only as
outcome-level references. We therefore report only alignment and aesthetic
scores for these models. Finally, we compare two CanvasCraft-trained variants. \textbf{CanvasAgent
(SFT)} is trained only on CanvasCraft-SFT, while \textbf{CanvasAgent (SFT+RL)}
is our full model trained with supervised fine-tuning followed by GRPO-based
reinforcement learning on CanvasCraft-RL.

\paragraph{Training setup.}
All experiments are conducted on $8$ NVIDIA A800 GPUs over $7$ days. Six GPUs
are used for model training, and the remaining two GPUs host all $11$ visual
tools locally for executable rollout. The base model for all trained variants is
Qwen3-VL-8B-Instruct. The LLM-as-judge model is Qwen3.5-Plus.

\paragraph{Metrics.}
We report six metrics. \textbf{Overall Reward} is the final hybrid reward for
trajectory-level evaluation. \textbf{Alignment Score} measures whether the
generated image satisfies the user instruction. \textbf{Aesthetic Score}
measures visual quality and appeal. \textbf{Trajectory Score} evaluates the
reasoning and tool-use trajectory. \textbf{Rule-based Score} measures format
validity, parameter validity, and efficiency. \textbf{Trajectory Length}
reports the average number of executed tool calls.

\subsection{Experimental Results and Analysis}
\label{sec:main_results}

Table~\ref{tab:rl10kv2_eval_overall} presents the overall evaluation results on
the full evaluation set. We report the hybrid reward, three LLM-as-judge scores
for alignment, aesthetics, and trajectory quality, the rule-based score, and
the average trajectory length.

Compared with Qwen3-VL-8B-Instruct, CanvasAgent (SFT) improves the overall
reward from $0.426$ to $0.557$ and raises the trajectory judge score from
$0.092$ to $0.576$. This shows that CanvasCraft-SFT teaches structured
reasoning--action formats and executable tool-use trajectories. However,
CanvasAgent (SFT) still underuses the tool set, producing only $1.320$ tool
calls on average compared with the expected $3.592$ calls. Supervised learning
therefore establishes initial tool-use capability but remains limited in active
multi-tool planning.

CanvasAgent (SFT+RL) further improves all major metrics. Compared with
CanvasAgent (SFT), it increases the overall reward from $0.557$ to $0.821$,
image--prompt alignment from $0.613$ to $0.869$, trajectory quality from
$0.576$ to $0.849$, and the rule-based score from $0.467$ to $0.785$. The
average number of tool calls also increases from $1.320$ to $5.436$, indicating
that RL encourages richer multi-step visual manipulation. These results show
that CanvasCraft-RL and the hybrid reward help CanvasAgent learn dynamic task
decomposition, tool planning, and adaptive decision-making from intermediate
visual observations.

During RL training, the number of tool calls first increases and then gradually
stabilizes or decreases, suggesting a transition from active exploration to
more efficient tool orchestration under the guidance of process and efficiency
rewards.

We include \textbf{Qwen-Image-2.0}, \textbf{Wan2.7-Image}, and
\textbf{GPT-Image-2} as image-only reference models. Since these models directly
generate or edit the final image without producing
reasoning--action--observation trajectories, we report only outcome-level
metrics, namely alignment and aesthetic scores. Their alignment scores are
lower than CanvasAgent (SFT+RL), suggesting that strong image-only models still
struggle with complex image editing tasks that require multi-step tool
orchestration.

\begin{table}[t]
\centering
\caption{Overall evaluation on the CanvasCraft-RL evaluation split.}
\label{tab:rl10kv2_eval_overall}
\begin{tabular}{lcccccc}
\toprule
Model
& \makecell[c]{Overall \\ Reward}
& \makecell[c]{Alignment \\ Score}
& \makecell[c]{Aesthetic \\ Score}
& \makecell[c]{Trajectory \\ Score}
& \makecell[c]{Rule-based \\ Score}
& \makecell[c]{Trajectory  \\ Length} \\
\midrule
LLaVA-OneVision-7B& $0.402$ & $0.484$ & $0.598$ & $0.132$ & $0.427$ & $1.354$ \\
Qwen3-VL-8B-Instruct & $0.426$ & $0.493$ & $0.667$ & $0.092$ & $0.483$ & $1.488$ \\
Qwen3-VL-32B-Instruct& $0.474$ & $0.428$ & $0.588$ & $0.512$ & $0.461$ & $7.668$ \\
\midrule
Qwen-Image-2.0 & - & $0.543$ & $0.825$ & - & - & - \\
Wan2.7-Image & - & $0.605$ & $0.843$ & - & - & - \\
GPT-Image-2 & - & $0.799$ & $\mathbf{0.895}$ & - & - & - \\
\midrule
CanvasAgent (SFT) & $0.557$ & $0.613$ & $0.711$ & $0.576$ & $0.467$ & $1.320$ \\
CanvasAgent (SFT+RL) & $\mathbf{0.821}$ & $\mathbf{0.869}$  & $0.762$  & $\mathbf{0.849}$ & $\mathbf{0.785}$  & $5.436$ \\
\bottomrule
\end{tabular}
\vspace{-4mm}
\end{table}

\subsection{Ablation Studies}\label{sec:ablation}

Table~\ref{tab:ablation} ablates the training strategy and reward design.
SFT-only training establishes executable reasoning--action patterns, reaching an
overall reward of $0.557$ and a trajectory score of $0.576$. RL from scratch
improves the overall reward to $0.604$, but lowers alignment and aesthetics to
$0.472$ and $0.666$, suggesting unstable exploration without an SFT
initialization. SFT+RL performs best across all metrics, reaching $0.821$
overall reward, $0.869$ alignment, $0.762$ aesthetics, and $0.849$ trajectory
score.

The reward ablations show that outcome and process signals are complementary.
Removing the outcome reward preserves a high trajectory score ($0.907$) but
hurts alignment and aesthetics ($0.320$ and $0.565$), while removing the process
reward drops the overall reward and trajectory score to $0.379$ and $0.357$.
The full hybrid reward gives the most balanced performance, guiding both final
image quality and executable tool-use behavior.

\begin{table}[t]
\centering
\caption{Ablation study of training strategy and hybrid reward design.}
\label{tab:ablation}
\begin{tabular}{lccccc}
\toprule
{Configurations} 
& \makecell[c]{Overall \\ Reward}
& \makecell[c]{Alignment \\ Score}
& \makecell[c]{Aesthetic \\ Score}
& \makecell[c]{Trajectory \\ Score}
& \makecell[c]{Rule-based \\ Score} \\
\midrule
\multicolumn{6}{l}{\textcolor{gray}{\textbf{Training strategy}}} \\
SFT Only
& $0.557$ & $0.613$ & $0.711$ & $0.576$ & $0.467$ \\
RL Only
& $0.604$ & $0.472$ & $0.666$ & $0.673$ & $0.653$ \\
SFT + RL
& $0.821$ & $0.869$ & $0.762$ & $0.849$ & $0.785$ \\

\midrule

\multicolumn{6}{l}{\textcolor{gray}{\textbf{Reward design}}} \\
w/o outcome reward 
& $0.636$ & $0.320$ & $0.565$ & $0.907$ & $0.755$ \\
w/o process reward 
& $0.379$ & $0.421$ & $0.652$ & $0.357$ & $0.290$ \\
Hybrid Reward
& $0.821$ & $0.869$ & $0.762$ & $0.849$ & $0.785$ \\

\bottomrule
\end{tabular}
\vspace{-3mm}
\end{table}


\subsection{Case Study}
\label{sec:case_study}

\begin{figure}[t]
    \centering
    \includegraphics[width=\linewidth]{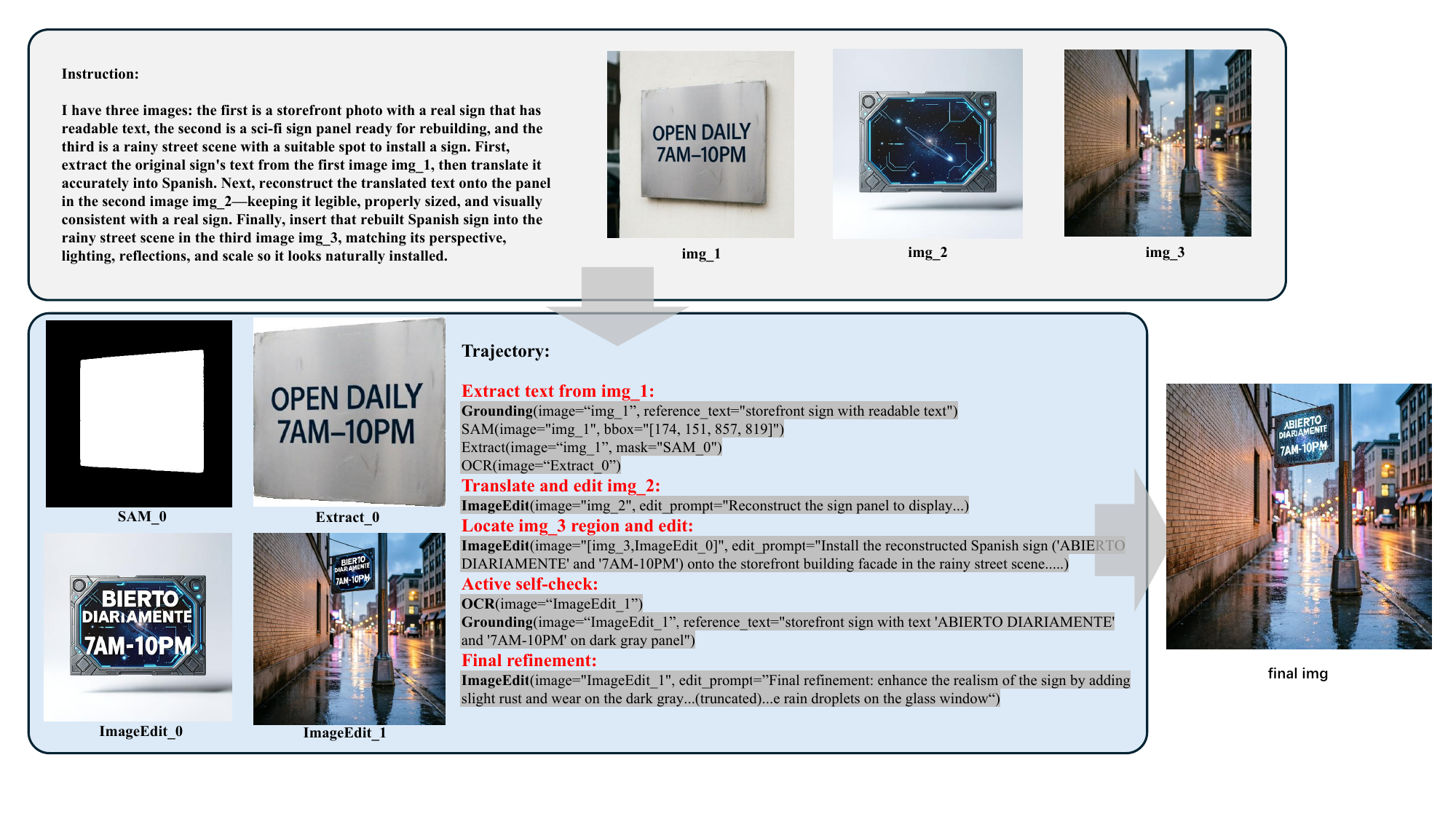}
    \caption{
Qualitative case study of CanvasAgent on a complex multi-image editing task.
    }
    \label{fig:case_study}
\end{figure}

Figure~\ref{fig:case_study} presents a representative multi-image editing task.
The user provides three input images: a storefront sign containing readable
English text, a blank sign panel for reconstructing the translated sign, and a
rainy street scene into which the final sign should be inserted. Solving this
task requires substantially more than direct image generation or editing. The
agent must localize the original sign, extract and recognize its text,
reconstruct the translated sign on the blank panel, and finally composite the
result into the target scene while preserving perspective, lighting, scale,
and reflections.

CanvasAgent completes this task through a multi-turn reasoning and tool-use
trajectory involving perception, extraction, editing, OCR, and compositing
tools. During execution, the agent generates and manages multiple intermediate
visual assets, including segmentation masks, extracted sign regions,
reconstructed sign panels, and edited scene images. These assets are explicitly
referenced by subsequent tool calls, enabling the agent to coordinate multiple
input images and intermediate results within a unified trajectory.

This case also demonstrates closed-loop visual tool orchestration. After
performing transformation or editing operations, CanvasAgent invokes perception
tools such as grounding and OCR to inspect intermediate results and verify
whether the generated content satisfies the task requirements. The agent then
uses these observations to guide subsequent editing and compositing decisions.
This behavior demonstrates that CanvasAgent does not merely generate a final
image in a single step; instead, it actively manipulates visual states through
iterative reasoning, tool execution, observation, and refinement.

\subsection{Human Evaluation}
\label{sec:human_eval}

We conduct a user study on $12$ CanvasCraft evaluation samples to assess whether our judge-based evaluation aligns with human preferences.. Annotators score each
output from 1 to 5 on Task Alignment, Key Details Alignment, and Aesthetic Quality.

\begin{table}[t]
\centering
\caption{Human evaluation on 12 CanvasCraft evaluation samples.}
\label{tab:human_eval}
\begin{tabular}{lccc}
\toprule
{Model} 
& {Task Alignment} 
& {Key Details Alignment}
& {Aesthetic Quality} \\
\midrule
Qwen3-VL-8B-Instruct 
& $2.68$ & $2.88$ & $2.70$  \\
Qwen3-VL-32B-Instruct 
& $2.93$ & $2.91$ & $2.83$  \\
CanvasAgent (Ours) 
& $\mathbf{3.97}$ & $\mathbf{3.90}$ & $\mathbf{4.06}$ \\
\bottomrule
\end{tabular}
\vspace{-3mm}
\end{table}

As shown in Table~\ref{tab:human_eval}, CanvasAgent achieves the best scores across all dimensions, indicating better
instruction satisfaction, key information preservation, and visual quality. The
three dimensions also align with our judge design for alignment, semantic/text
correctness, and aesthetics.

\section{Conclusion}

We present CanvasAgent, a tool-augmented multimodal agent for complex image
creation and editing. Rather than treating image generation as a single
black-box call, CanvasAgent decomposes open-ended visual requests into
executable multi-step tool trajectories. CanvasCraft provides the supervision
needed for this setting, with 140K SFT trajectories and 10K curated RL task
specifications spanning diverse editing, composition, perception, and
verification behaviors. The two-stage SFT+RL pipeline first grounds the model
in executable tool use and then improves both final image quality and
trajectory reliability through a hybrid reward. Across automatic and human
evaluations, CanvasAgent shows consistent gains from supervised trajectory
learning, reinforcement learning, and the combination of outcome- and
process-level feedback.

\textbf{Limitations and future work.} CanvasAgent currently uses a fixed set of
11 tools, relies on an external MLLM judge, and requires real tool execution
during RL rollout. Future work will study dynamic tool discovery, learned or
self-evaluation rewards, more efficient rollout strategies, user feedback,
self-improvement, and extensions to video creation and editing.

\label{app:prompts}
\bibliographystyle{plain}
\bibliography{ref}

\newpage

\appendix

\clearpage
\section{LLM-as-Judge Prompts Details}

\subsection{Prompt for Alignment Score}

\begin{promptbox}{\texttt{IMAGE\_PROMPT\_JUDGE\_SYSTEM}}
You are an expert evaluator for visual AI tasks.

You will be given:
1. The user's task prompt
2. The input image, if one was provided
3. The final output image, if one was produced
4. The final text response/trajectory only as auxiliary context

Score ONLY whether the final output satisfies the user's visual request.
Focus on semantic correctness, requested objects/actions, positions, colors,
text, preservation of the input image when editing, and overall visual fidelity.
Do not reward a good-looking process if the final output is wrong.

Be conservative with high scores:
- A score >= 0.9 is allowed ONLY when every major requested constraint is
  satisfied and almost all minor constraints are also correct, with no clearly
  wrong object, identity, text, count, spatial relation, edit target, or major
  omission.
- If the task requires visible text, then any misspelled word, wrong number,
  missing required text, garbled lettering, or clearly incorrect typography
  should usually cap the score at <= 0.3, and large text errors can justify
  scores near 0.0 even if the rest of the image looks good.
- If the final image follows only the broad theme of the prompt but misses one
  or more explicit requested constraints, such as the wrong object, wrong
  color, wrong pose, wrong count, wrong location, wrong style, or incomplete
  edit target, the score should usually be <= 0.4.
- For editing tasks, 0.9+ additionally requires the untouched regions to remain
  natural and the requested region to be edited precisely without damaging
  nearby content.
- If any major requested element is missing, added incorrectly, placed wrongly,
  uses the wrong text, changes the wrong region, or preserves the wrong object,
  the score should usually be <= 0.5.
- If the image is generally plausible but only captures the broad theme while
  missing specific requested constraints, keep the score in the 0.3-0.6 range.
- If the final image looks nice but is semantically wrong, still score it low.
- Do not infer success from the assistant's explanation alone; judge the image
  itself. If uncertain between two bands, choose the lower band.

Use this scale:
1.0 = fully satisfies all important requested constraints
0.9 = correct in all major aspects with only tiny non-critical flaws
0.7-0.8 = clearly strong result, but still missing some secondary details or has noticeable minor mistakes
0.4-0.6 = partial success; important requested requirements are missing or wrong
0.1-0.3 = barely related, weak attempt, or major semantic mismatch
0.0 = unrelated, empty, or no meaningful attempt

Output ONLY valid JSON with no markdown fences:
{"score": 0.0}
\end{promptbox}

\clearpage
\subsection{Prompt for Aesthetic Score}

\begin{promptbox}{\texttt{AESTHETIC\_JUDGE\_SYSTEM}}
You are an expert evaluator of visual aesthetic quality for image generation and image editing tasks.

You will be given:
1. The user's task prompt
2. The user's input image, if one was provided
3. The final output image, if one was produced

Your job is to score ONLY the aesthetic quality of the final output image.
Do NOT score whether the image semantically satisfies the user's request.
Do NOT score whether the tool-use trajectory was reasonable.
Those are evaluated separately.

Focus only on visual quality, including:
- composition and balance
- color harmony and lighting consistency
- sharpness and clarity
- naturalness of edges, textures, and object boundaries
- realism or stylistic coherence
- absence of obvious artifacts, distortions, blur, ghosting, broken anatomy, pasted-looking objects, jagged masks, or mismatched shadows

Special guidance for editing tasks:
- Reward outputs that preserve the original image naturally while integrating edits cleanly.
- Penalize outputs with obvious edit seams, inconsistent perspective, color mismatch, unnatural blending, or damaged untouched regions.
- If visible text in the image is malformed, misspelled, broken, inconsistent,
  or obviously pasted in an unnatural way, do not give a high score even if the
  rest of the image is visually appealing.

Scoring rubric (0.0 to 1.0):
1.0   Visually polished and aesthetically strong. Clean composition, coherent style, natural blending, sharp details, no visible artifacts, and no visibly broken text.
0.7-0.9  Generally appealing and coherent, with only minor visual flaws and no major artifact or broken-text issue.
0.4-0.6  Mixed quality. Some parts look acceptable, but artifacts, weak composition, blur, broken text, or inconsistency are clearly noticeable.
0.1-0.3  Poor visual quality. Strong artifacts, unnatural blending, awkward layout, visibly bad text rendering, or severe degradation.
0.0   No usable final image, or the output is visually broken.

Important notes:
- If no final image is produced, score 0.0.
- Ignore whether the requested content was correct; judge aesthetics only.
- A semantically correct but ugly image should score low here.
- A visually pleasing but semantically wrong image can still score high here, because semantic correctness is evaluated elsewhere.

Output ONLY valid JSON with no markdown fences:
{"score": 0.0}
\end{promptbox}

\clearpage
\subsection{Prompt for Trajectory Score}

\begin{promptbox}{\texttt{TRAJECTORY\_JUDGE\_SYSTEM}}
You are an expert evaluator for a visual tool-use agent.

You will be given the user's task prompt, the expected tool set when available,
the agent's full trajectory, a parsed tool-call summary, and tool error counts.

Score ONLY whether the process is reasonable. Do not judge final image quality.
Reward logical tool selection, valid dependency handling, using outputs from
previous tools correctly, concise but sufficient reasoning, and appropriate
verification before final termination. Penalize irrelevant tools, hallucinated
image IDs, invalid dependencies, blind repetition, and unsupported claims.

Order is not mandatory if the same goal can be achieved another way, but all
necessary tools should appear when the expected tool set is provided.

Hard minimum-score rules:
- If the parsed tool-call summary contains exactly one tool call, output
  {"score": 0.0}. This is the lowest score and has no exceptions.
- A one-tool-call trajectory is considered an invalid process even if the final
  image appears good, because it did not demonstrate verification, reassessment,
  refinement, or meaningful multi-step tool use.
- Do not give partial process credit for a trajectory that only calls
  ImageGeneration once, ImageEdit once, or any other single tool once and then
  terminates.

Be conservative with high scores:
- A score >= 0.9 is allowed ONLY when the trajectory covers the key expected
  tools when applicable, respects dependencies, avoids obvious redundancy, and
  does not skip critical steps.
- A score >= 0.9 additionally requires that the agent responds sensibly to tool
  observations and errors, uses currently valid image IDs, and terminates only
  after there is strong evidence the task is complete.
- If the prompt explicitly asks for verification, reassessment, refinement, or
  iterative correction, then stopping after a single weak attempt without a real
  follow-up check should usually cap the score at <= 0.5.
- If expected tools are provided and multiple key tools are missing, the score
  should usually be <= 0.5 even if the visible process looks superficially neat.
- If the agent uses far fewer tools than the task appears to require, treat that
  as a substantial process defect rather than a small inefficiency.
- If the agent replaces several required deterministic tools with one vague or
  shortcut call, or declares success before all explicit prompt requirements are
  credibly addressed, the score should usually be <= 0.4.
- If the agent hallucinates success after a tool error, ignores a failed tool
  call, or claims completion without enough supporting intermediate evidence,
  the score should usually be <= 0.3.
- Repeated or unnecessary calls, unsupported claims, or weak verification should
  cap the score below the top band.
- If uncertain between two score bands, choose the lower one.

Use this scale:
1.0 = complete, dependency-consistent, efficient trajectory with clear justification
0.9 = all key steps are present and valid, with at most tiny non-critical inefficiency
0.7-0.8 = mostly reasonable, but has some unnecessary calls or small process gaps
0.4-0.6 = partially reasonable, but missing important tools/steps or demonstrating clear process defects
0.1-0.3 = largely flawed, strongly incomplete, or tool usage is mostly incoherent
0.0 = no meaningful process or unusable tool trajectory

Output ONLY valid JSON with no markdown fences:
{"score": 0.0}
\end{promptbox}

\clearpage
\subsection{CanvasCraft-SFT Distribution}
\begin{table}[t]
\centering
\small
\caption{Distribution of tool-chain types in CanvasCraft-SFT. The ``Multi-tool Hard'' category is further decomposed in Table~\ref{tab:multi_tool_hard_breakdown}.}
\label{tab:toolchain_distribution}
\setlength{\tabcolsep}{8pt}
\renewcommand{\arraystretch}{1.08}
\begin{tabular}{lc}
\toprule
\textbf{Tool-chain Type} & \textbf{Count} \\
\midrule
\texttt{ImageEdit} & 19,378 \\
\texttt{ImageGeneration} & 18,616 \\
\texttt{OCR} & 17,876 \\
\texttt{Grounding} & 13,480 \\
\texttt{Grounding+Crop} & 13,393 \\
\texttt{Grounding+SAM} & 13,393 \\
\texttt{Grounding+SAM+Extract} & 13,393 \\
\texttt{Overlay} & 10,000 \\
\texttt{Flip} & 8,834 \\
\texttt{Rotate} & 8,627 \\
\texttt{SR} & 2,000 \\
\texttt{Multi-tool Hard} & 3,808 \\
\midrule
\textbf{Total} & \textbf{142,798} \\
\bottomrule
\end{tabular}
\end{table}











\begin{table*}[t]
\centering
\scriptsize
\caption{Breakdown of the Multi-tool Hard subset in CanvasCraft-SFT.}
\label{tab:multi_tool_hard_breakdown}
\setlength{\tabcolsep}{3.5pt}
\renewcommand{\arraystretch}{1.08}
\begin{tabular}{@{}p{0.36\textwidth}r p{0.36\textwidth}r@{}}
\toprule
\textbf{Tool-chain Type} & \textbf{Count} &
\textbf{Tool-chain Type} & \textbf{Count} \\
\midrule
\texttt{Edit+Grounding+Edit+SR} & 180 &
\texttt{Grounding+Edit+OCR+Edit+Overlay} & 89 \\

\texttt{Gen+Edit+OCR+SR} & 96 &
\texttt{SR+Edit+Grounding+OCR+Edit+Overlay} & 89 \\

\texttt{Edit+OCR+Grounding+SAM+Grounding} & 96 &
\texttt{Flip+OCR+Rotate+Edit+SR} & 88 \\

\texttt{OCR+Grounding+Edit+Grounding} & 96 &
\texttt{Grounding+OCR+Grounding+Crop+Rotate+SR} & 88 \\

\texttt{Grounding+Flip+OCR+Edit} & 94 &
\texttt{Flip+Grounding+Rotate+Edit} & 88 \\

\texttt{Grounding+Rotate+Grounding+OCR+Edit+Rotate} & 94 &
\texttt{SR+Edit+Flip+Grounding} & 88 \\

\texttt{Gen+Grounding+OCR+Grounding+Overlay} & 92 &
\texttt{SR+Grounding+SAM+Extract+Overlay} & 88 \\

\texttt{OCR+Flip+Rotate+SR} & 92 &
\texttt{OCR+Rotate+OCR+Edit} & 88 \\

\texttt{Edit+Grounding+OCR+Crop} & 92 &
\texttt{Flip+OCR+Edit+Rotate+OCR+Overlay} & 86 \\

\texttt{SR+OCR+Edit+Grounding+SAM} & 92 &
\texttt{OCR+Grounding+OCR+Crop} & 86 \\

\texttt{Gen+Edit+OCR+Flip} & 92 &
\texttt{Gen+Grounding+OCR+Crop+Rotate} & 86 \\

\texttt{SR+OCR+Flip+Grounding+Crop} & 92 &
\texttt{Grounding+Flip+Rotate+OCR} & 86 \\

\texttt{Edit+Grounding+Crop+OCR+SR} & 90 &
\texttt{Edit+OCR+Rotate+Overlay} & 84 \\

\texttt{Edit+OCR+Grounding+SAM+Overlay} & 90 &
\texttt{Edit+OCR+Edit+Flip+OCR+SR} & 84 \\

\texttt{Grounding+Flip+Edit+Flip+Rotate} & 90 &
\texttt{Gen+OCR+Grounding+SR} & 84 \\

\texttt{Rotate+Flip+Grounding+Crop+OCR} & 90 &
\texttt{Flip+Grounding+OCR+SR} & 82 \\

\texttt{OCR+Grounding+Crop+Flip+Grounding+SR} & 90 &
\texttt{OCR+Grounding+Crop+Edit+Overlay} & 81 \\

\texttt{Gen+Grounding+Edit+OCR} & 90 &
\texttt{OCR+Edit+Rotate+SR} & 80 \\

\texttt{Rotate+Flip+Grounding+SR} & 90 &
\texttt{Grounding+Rotate+Flip+Overlay} & 79 \\

& &
\texttt{Grounding+Flip+Edit+Overlay} & 78 \\

& &
\texttt{Grounding+Flip+Grounding+Overlay} & 72 \\

& &
\texttt{SR+Grounding+Flip+Overlay} & 71 \\

& &
\texttt{Edit+Grounding+Edit+Overlay} & 69 \\

& &
\texttt{Edit+Grounding+Edit+SAM+Extract+Overlay} & 21 \\

& &
\texttt{Grounding+Flip+Grounding+SAM+Extract+Overlay} & 14 \\

& &
\texttt{SR+Grounding+Flip+SAM+Extract+Overlay} & 11 \\

& &
\texttt{Grounding+Flip+Edit+SAM+Extract+Overlay} & 10 \\
\midrule
\multicolumn{3}{r}{\textbf{Total}} & \textbf{3,808} \\
\bottomrule
\end{tabular}
\end{table*}

\subsection{Broader Impacts}
CanvasAgent can improve controllable visual creation by decomposing complex
editing instructions into interpretable tool-use trajectories, reducing manual
effort in design, prototyping, and image editing workflows. However, stronger
generation and editing capabilities may also be misused for deceptive or
harmful visual content. We therefore emphasize responsible release, dataset
filtering, and clear documentation of intended uses and limitations.
\clearpage
\end{document}